\documentclass[final,journal,twocolumn]{IEEEtran}

\usepackage[pdftex]{graphicx}
\usepackage{amsmath}
\usepackage{amssymb}
\usepackage{booktabs}
\usepackage{tabularx}
\usepackage{makecell}

\DeclareMathOperator*{\argmax}{\arg\!\max}

\usepackage[noend]{algpseudocode}
\usepackage{algorithm}

\usepackage{hyperref}
\hypersetup{
    colorlinks=true,
    linkcolor=black,
    citecolor=black,
    urlcolor=blue,
}
\usepackage{multirow}

\usepackage{cite}
\usepackage{tikz}
\usepackage{pgfplots}
\usepgfplotslibrary{colorbrewer}
\pgfplotsset{cycle list/Set1-9}
\tikzset{every picture/.style={line width=1pt}}
\usetikzlibrary{patterns}
\usepgfplotslibrary{fillbetween}
\usepackage[eulergreek]{sansmath}
\pgfplotsset{
  tick label style = {font=\sansmath\sffamily\scriptsize},
  every axis label = {font=\sansmath\sffamily\scriptsize},
  y label style={at={(0.05,0.5)}},
  legend style = {font=\sansmath\sffamily\scriptsize},
  label style = {font=\sansmath\sffamily\scriptsize},
}
\pgfplotsset{compat=1.17}
\DeclareMathAlphabet\mathbfcal{OMS}{cmsy}{b}{n}
\newlength{\Oldarrayrulewidth}
\newcommand{\Cline}[2]{%
  \noalign{\global\setlength{\Oldarrayrulewidth}{\arrayrulewidth}}%
  \noalign{\global\setlength{\arrayrulewidth}{#1}}\cline{#2}%
  \noalign{\global\setlength{\arrayrulewidth}{\Oldarrayrulewidth}}}

\usepackage[most]{tcolorbox}
\tcbset{top=0.01mm,left=0.01mm,bottom=0.01mm,right=0.01mm}
\usepackage{efbox}
\efboxsetup{linecolor=green,linewidth=10pt}
\usepackage{color, soul}

\makeatletter
\def\BState{\State\hskip-\ALG@thistlm}
\makeatother

\newcommand{\signscale}{0.76}
\newcommand{\signraise}{0.24\height}
\newcommand{\minus}{\raisebox{\signraise}{\scalebox{\signscale}{$-$}}}
\newcommand{\plus}{\raisebox{\signraise}{\scalebox{\signscale}{$+$}}}
\newcommand{\product}{\raisebox{\signraise}{\scalebox{\signscale}{$\times$}}}
\newcommand{\equal}{\raisebox{\signraise}{\scalebox{\signscale}{$=$}}}

\begin{document}

\title{Informative and Representative Triplet Selection for Multilabel Remote Sensing Image Retrieval}
 
\author{Gencer~Sumbul,~\IEEEmembership{Graduate Student Member,~IEEE}
        Mahdyar~Ravanbakhsh,~\IEEEmembership{Member,~IEEE,}
        and~Begüm~Demir,~\IEEEmembership{Senior~Member,~IEEE}
\thanks{Gencer Sumbul, Mahdyar Ravanbakhsh and Begüm Demir are with the Faculty of Electrical Engineering and Computer Science, Technische Universit\"at Berlin, 10623 Berlin, Germany. Email: \mbox{gencer.suembuel@tu-berlin.de}, \mbox{ravanbakhsh@tu-berlin.de}, \mbox{demir@tu-berlin.de}.}}


\maketitle

\begin{abstract}
Learning the similarity between remote sensing (RS) images forms the foundation for content-based RS image retrieval (CBIR). Recently, deep metric learning approaches that map the semantic similarity of images into an embedding (metric) space have been found very popular in RS. A common approach for learning the metric space relies on the selection of triplets of similar (positive) and dissimilar (negative) images to a reference image called as an anchor. Choosing triplets is a difficult task particularly for multi-label RS CBIR, where each training image is annotated by multiple class labels. To address this problem, in this paper we propose a novel triplet sampling method in the framework of deep neural networks (DNNs) defined for multi-label RS CBIR problems. The proposed method selects a small set of the most representative and informative triplets based on two main steps. In the first step, a set of anchors that are diverse to each other in the embedding space is selected from the current mini-batch using an iterative algorithm. In the second step, different sets of positive and negative images are chosen for each anchor by evaluating the relevancy, hardness and diversity of the images among each other based on a novel strategy. Experimental results obtained on two multi-label benchmark archives show that the selection of the most informative and representative triplets in the context of DNNs results in: i) reducing the computational complexity of the training phase of the DNNs without any significant loss on the performance; and ii) an increase in learning speed since informative triplets allow fast convergence. The code of the proposed method is publicly available at \url{https://git.tu-berlin.de/rsim/image-retrieval-from-triplets}.
\end{abstract}


\IEEEpeerreviewmaketitle

\section{Introduction}
\IEEEPARstart{I}{n} recent years, advancements in satellite technology have led to fast-growing archives of remote sensing (RS) images. One of the most emerging applications in RS is the accurate retrieval of RS images from such archives. Thus, the development of content-based image retrieval (CBIR) methods has recently attracted great attention \cite{rs-book}. The performance of any CBIR system relies on its capability to learn discriminative and robust image representations to describe the complex semantic content of RS images.

Conventional CBIR systems exploit hand-crafted features to describe the content of images. As an example, Wang and Newsam present a retrieval system employing the well-known scale-invariant feature transform (SIFT) to extract bag-of-visual-words representations of image features \cite{bovw-sift}. Aptoula introduces the use of bag-of-morphological-words representations for local texture descriptors \cite{bomw-morph}. In \cite{study-lbps}, a comparative analysis of local binary patterns (LBP) that capture local patterns between neighboring pixels is presented. Chaudhuri et al. present a method that represents image content by a graph, where the graph nodes describe the image region properties and the edges represent the spatial relationships among the regions \cite{ucmerced-multi}. Binary hash codes obtained through kernel-based hashing methods are found effective for describing RS images in \cite{hashing-retrieval}. After extracting the image features, the most similar images with respect to a query image can be found by performing the $k$-nearest neighbor ($k$-nn) search algorithm. In the case of graph-based image representations, graph comparison methods such as the inexact graph matching approach proposed by Chaudhuri et al. \cite{region-graph} can be used. The images represented by binary hash codes can be searched and retrieved by using the computationally efficient hamming distance \cite{hashing-retrieval}.

The above-mentioned CBIR systems cannot simultaneously optimize feature learning and image retrieval, and thus result in a limited capability to represent the high-level semantic content of RS images. This issue leads to insufficient search and retrieval performance \cite{rs-book}. To overcome this problem, CBIR systems based on deep neural networks (DNNs) have been recently presented in RS \cite{related-simul-hashing}. As an example, Li et al. propose a method that fuses deep features and hand-crafted features \cite{li2016cbrsir}. This method exploits four convolutional neural networks (CNNs) to extract features at different steps and with different coarse levels. Then, these deep features are fused with traditional image descriptors such as LBPs and SIFT to be used in the retrieval process. A convolutional autoencoder is used by Tang et al. to obtain deep bag-of-words image descriptors in \cite{tang2018unsupervised}. To this end, a reconstruction loss function that minimizes the error between the input and the extracted descriptors is considered. Imbriaco et al. extract local convolutional features and aggregate them into a global descriptor, where the deep features are extracted through a pre-trained model without any fine-tuning~\cite{imbriaco2019aggregated}. Boualleg and Farrah address the semantic gap between low-level features and high-level perception of semantic similarity in \cite{boualleg2018enhanced}. This is achieved by using a CNN to detect semantic concepts and a relevance feedback strategy to ensure that CBIR results match with a query image. Sabahi et al. address the above-mentioned semantic gap by employing a recurrent neural network to model the human visual memory \cite{related-hopfield}.

\begin{figure}[t]
    \newcommand{\figwidth}{0.31\linewidth} 
    \newcommand{\captionwidth}{0.9\linewidth}
    \newcommand{\figheight}{0.63in} 
    \renewcommand{\fboxsep}{0pt}
    \centering
    \scriptsize
   \begin{minipage}[t]{\figwidth}
        \centering
        \centerline{\tcbox[sharp corners,colframe=blue!70!black, top=-1pt,left=-1pt,right=-1pt,bottom=-1pt]{\includegraphics[height=\figheight]{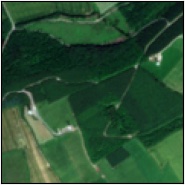}}}
        \vspace{0.03in}
        \centerline{\parbox{\captionwidth}{\centering Arable land, Pastures,\\Coniferous forest}}
        \vspace{0.2in}
    \end{minipage}
    \hspace{0.05in}
     \begin{minipage}[t]{\figwidth}
        \centering
        \centerline{\tcbox[sharp corners,colframe=green!70!black, top=-1pt,left=-1pt,right=-1pt,bottom=-1pt]{\includegraphics[height=\figheight]{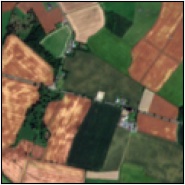}}}
        \vspace{0.03in}
        \centerline{\parbox{\captionwidth}{\centering Arable land,\\Pastures}}
    \end{minipage}
    \hspace{0.05in}
     \begin{minipage}[t]{\figwidth}
        \centering
        \centerline{\tcbox[sharp corners,colframe=red!70!black, top=-1pt,left=-1pt,right=-1pt,bottom=-1pt]{\includegraphics[height=\figheight]{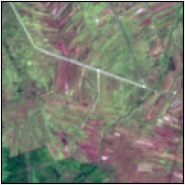}}}
        \vspace{0.03in}
        \centerline{\parbox{\captionwidth}{\centering Mixed forest,\\Inland wetlands}}
    \end{minipage}
    \vspace{0.1in}
       \begin{minipage}[t]{\figwidth}
        \centering
        \centerline{\tcbox[sharp corners,colframe=blue!70!black, top=-1pt,left=-1pt,right=-1pt,bottom=-1pt]{\includegraphics[height=\figheight]{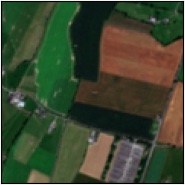}}}
        \vspace{0.03in}
        \centerline{\parbox{\captionwidth}{\centering Arable land, Pastures,\\Complex cultivation patterns}}
    \end{minipage}
    \hspace{0.05in}
     \begin{minipage}[t]{\figwidth}
        \centering
        \centerline{\tcbox[sharp corners,colframe=green!70!black, top=-1pt,left=-1pt,right=-1pt,bottom=-1pt]{\includegraphics[height=\figheight]{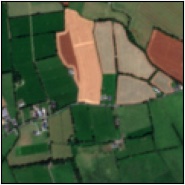}}}
        \vspace{0.03in}
        \centerline{\parbox{\captionwidth}{\centering Arable land, Pastures,\\Complex cultivation patterns}}
    \end{minipage}
    \hspace{0.05in}
     \begin{minipage}[t]{\figwidth}
        \centering
        \centerline{\tcbox[sharp corners,colframe=red!70!black, top=-1pt,left=-1pt,right=-1pt,bottom=-1pt]{\includegraphics[height=\figheight]{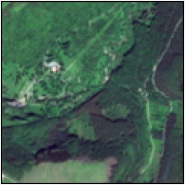}}}
        \vspace{0.03in}
        \centerline{\parbox{\captionwidth}{\centering Broad-leaved forest,\\Coniferous forest}}
    \end{minipage}
    \vspace{0.1in}
       \begin{minipage}[t]{\figwidth}
        \centering
        \centerline{\tcbox[sharp corners,colframe=blue!70!black, top=-1pt,left=-1pt,right=-1pt,bottom=-1pt]{\includegraphics[height=\figheight]{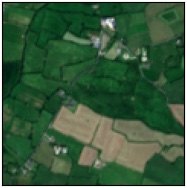}}}
        \vspace{0.03in}
        \centerline{\parbox{\captionwidth}{\centering Pastures,\\Coniferous forest}}
    \end{minipage}
    \hspace{0.05in}
     \begin{minipage}[t]{\figwidth}
        \centering
        \centerline{\tcbox[sharp corners,colframe=green!70!black, top=-1pt,left=-1pt,right=-1pt,bottom=-1pt]{\includegraphics[height=\figheight]{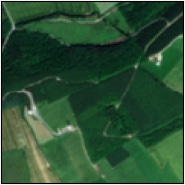}}}
        \vspace{0.03in}
        \centerline{\parbox{\captionwidth}{\centering Arable land, Pastures,\\Coniferous forest}}
    \end{minipage}
    \hspace{0.05in}
     \begin{minipage}[t]{\figwidth}
        \centering
        \centerline{\tcbox[sharp corners,colframe=red!70!black, top=-1pt,left=-1pt,right=-1pt,bottom=-1pt]{\includegraphics[height=\figheight]{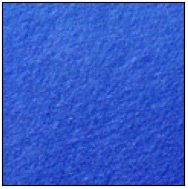}}}
        \vspace{0.03in}
        \centerline{\parbox{\captionwidth}{\centering Marine waters}}
    \end{minipage}
    \caption{An example of three triplets consisting of images from BigEarthNet \cite{bigearthnet}. Each triplet given in different rows consists of an anchor (in blue frame), a positive image (in green frame), and a negative image (in red frame). The associated multi-labels are given below the respective images.}
    \label{fig:triplet_samples}
\end{figure}

In recent years, deep metric learning (DML) based methods that aim at learning a feature space (in which similar images are close to each other) have attracted attention in RS. Current DML models are mostly trained using a triplet loss function made up of three images as: i) an \textit{anchor image}; ii) a \textit{positive image} that is similar to the anchor; and iii) a \textit{negative image} that is dissimilar to the anchor~\cite{triplet-loss}. An example of triplets constructed from BigEarthNet~\cite{bigearthnet} can be seen in Fig.~\ref{fig:triplet_samples}. A difficult task in DML is to construct the set of triplets. A simple strategy is to define triplets from an existing training set of labeled images. Roy et al. apply a strategy that: i) randomly selects an anchor from a mini-batch of training images; and then ii) randomly chooses one positive image that has the same class label as the anchor, while selecting one negative image that has a different class label \cite{metric-deep-hashing}. Similarly, Lai et al. select triplets randomly based on the class labels of training images to train an end-to-end model for hashing \cite{related-simul-hashing}. For each anchor image, there can be several positive and negative images. Thus, random selection does not guarantee the selection of the most representative and informative images to the anchor and can result in the construction of so-called \textit{trivial triplets} (see Section \ref{sec:related} for details). We would like to note that one can also exploit all the images in the mini-batch to construct triplets, as suggested in \cite{cao2020enhancing}. However, this choice significantly increases the total number of triplets and thus the computational complexity of the training phase of the retrieval system \cite{related-angular, person-reident}.

To overcome the limitation of random selection, the DML methods that evaluate the hardness of images during the sampling process are introduced in the computer vision (CV) literature (see Section \ref{sec:related} for details). According to our knowledge, most of the triplet sampling methods in CV assume that each image is annotated by a single label associated with the most significant content of the considered image and thus rely on single-label image annotations to decide which images are positive or negative for a given anchor image. However, RS images typically consist of multiple classes and thus can simultaneously be associated with different class labels (i.e., multi-labels). From the DML perspective, the selection of triplets from training images annotated by multi-labels is more complex than that from training images labeled by single-labels. To achieve accurate DML in multi-label RS CBIR, methods that accurately select a set of triplets from multi-label training images are needed.

To address this problem, we propose a novel triplet sampling method in the framework of DML designed for multi-label RS CBIR problems. Unlike the existing triplet sampling methods, the proposed method aims to select a small set of triplets from each mini-batch of multi-label training images. To this end, the proposed method consists of two consecutive steps. In the first step, a small number of diverse anchors is selected based on a simple but efficient iterative algorithm. In the second step, relevant, hard and diverse positive and negative images with respect to each anchor are chosen based on a novel strategy. Then, the triplets are constructed from the selected anchors and their respective positive and negative images. Based on these consecutive steps, the proposed method constructs a small number of the most informative and representative triplets to drive DML, resulting in an accurate CBIR and also in a reduced training complexity for the considered DNN. It is worth noting that the proposed triplet sampling method is independent of the considered DNN architecture, and therefore can be used within any DNN presented in the literature. In the experiments, different DNN architectures are considered, while the $k$-nn algorithm is used for the retrieval process after the characterization of the image descriptors through the considered method. Experiments carried out on two multi-label RS benchmark archives demonstrate the effectiveness of the proposed method. 

The rest of the paper is organized as follows: Section \ref{sec:related} presents the related works on triplet sampling. Section \ref{sec:proposed} introduces the proposed method. Section \ref{sec:setup} describes the considered datasets and the experimental setup, while Section \ref{sec:results} provides the experimental results. Section \ref{sec:conclusion} concludes our paper. 
\section{Related Works on Triplet Sampling}
\label{sec:related}
\begin{figure}[t]
  \centering
  \includegraphics[width=0.95\linewidth]{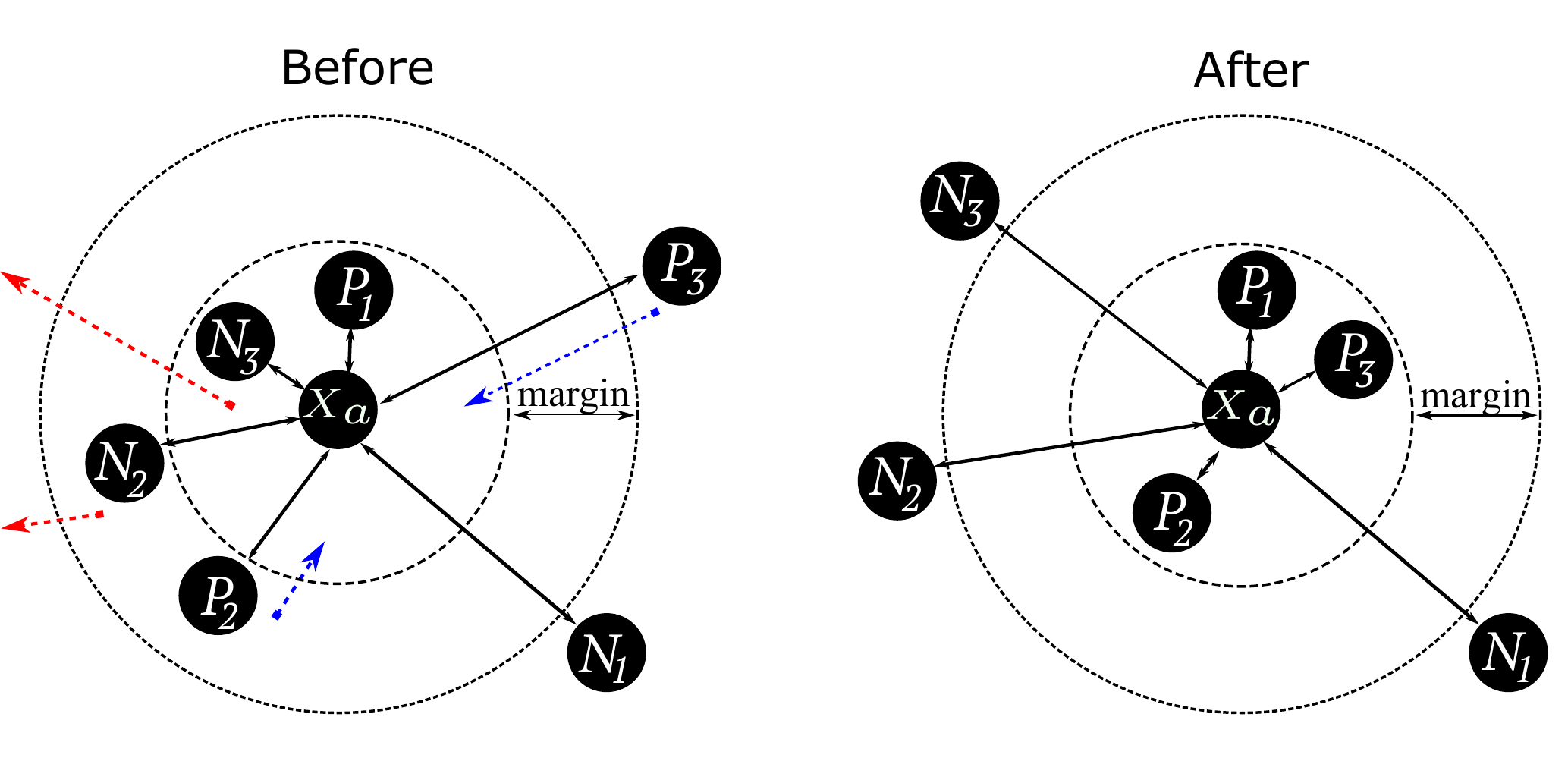}
  \caption{An Abstract representation of triplet selection and the progress for feature space update. Blue arrows indicate reducing distances for updating the embedding, while red arrows indicate increasing the distances. $X_a$ marks a chosen anchor, $P_1$, $P_2$, and $P_3$ are positive images, and $N_1$, $N_2$, and $N_3$ are negative images in different triplets. The triplet $(X_a, P_1, N_1)$ is trivial because it already satisfies the margins, and thus the corresponding distances are not updated. The triplet $(X_a, P_2, N_2)$ leads to a relatively small error and the images are pushed and pulled a little. The triplet $(X_a, P_3, N_3)$ violates the margin greatly and causes a significant error. $P_3$ is a positive image, but very far from the anchor, so it is considered as a hard positive image. $N_3$ is respectively a hard negative image.}
  \label{fig:triplet_scenes}
\end{figure}

The development of DML methods that aim to learn a metric space (in which semantically similar images are close to each other) is important for an accurate CBIR. It has been shown that the triplet-based DML methods perform considerably well for the CBIR tasks \cite{metric-deep-hashing, zhu2020deep}. The triplet-based DML methods use triplets of images to learn a metric space by means of the triplet loss \cite{triplet-loss}. The optimization objective is to minimize the feature distance between the anchor and its positive sample (i.e., image) while maximizing the feature distance between the anchor and the negative sample. The goal is to ensure that the positive sample is closer to the anchor than the negative sample by at least a margin. During the training of a triplet-based DML method, for the triplets that consist of a positive image inside the margin and the negative image outside the margin, a zero value triplet loss is obtained, leading to small gradient values and slow convergence. For the triplets that consist of a positive image visually less similar to the anchor (i.e., outside the margin) and a negative image visually more similar to the anchor (i.e., inside the margin), a high triplet loss value is obtained. High loss values lead to large gradient values, and thus the parameters of the model are updated. When a positive image is far from the margin, it is called as a \textit{hard positive} image. A negative image is called as \textit{hard negative} if it is inside the margin and very close to the anchor. If the distance between the anchor and positive image of a triplet is higher than the distance between the anchor and negative image, the triplet is considered as a \textit{hard triplet}. In Fig. \ref{fig:triplet_scenes}, an abstract representation of the triplet selection and the feature space update is demonstrated. The images $P_1$, $P_2$ and $P_3$ are the positive images for the anchor $X_a$ in different triplets, while images $N_1$, $N_2$ and $N_3$ are the negative images for the anchor $X_a$. After updating the embedding (metric) space using the selected triplets, $P_2$ and $P_3$ are pulled closer to the anchor $X_a$, while $N_2$ and $N_3$ pushed far away from the anchor $X_a$ towards outside the margin. The positive image $P_1$ is inside the margin and negative image $N_1$ is outside the margin, and thus triplet $(X_a,P_1,N_1)$ is a trivial triplet. The positive image $P_3$ is a hard positive image for anchor $X_a$, since it is outside the margin and far from the anchor image. The negative image $N_3$ is a hard negative image, as it is very close to the anchor. The triplet $(X_a,P_3,N_3)$ is a hard triplet, and causes a high loss value to update the parameters of the model. Since the trivial triplets are not sufficiently informative and lead to slow convergence, the use of hard triplets has been considered to overcome this problem. 

Most of the methods in RS do not consider the hardness of the images in the selected triplets and exploit the random triplet selection strategy as mentioned in the introduction \cite{metric-deep-hashing, cao2020enhancing,zhang2021triplet}. Unlike RS, in the CV community, the use of triplets is more extended and the importance of the hardness is widely studied \cite{xuan2020improved, zhang2020deep, ge2018deep, yuan2017hard}. As an example, Xuan et al. propose a triplet selection strategy that selects the closest positive sample (easy positive) and the closest negative (hard negative) for each anchor \cite{xuan2020improved}. Yuan et al. propose a hard-aware deeply cascaded (HDC) embedding method \cite{yuan2017hard}. For each anchor and a selected positive sample, HDC selects the negative samples at multiple hardness levels to construct different triplets. Hardness levels are defined based on the distances in the embedding space. Yang et al. investigate the importance of hard positive images by combining a positive image with all negative image pairs in the batch \cite{person-reident}. Then, the positive images are weighted and hard positives are preferred. Ge et al., propose a hard triplet selection method that constructs a class-level hierarchical tree of image features for the whole dataset, where visually similar classes are merged recursively \cite{ge2018deep}. Then, the selection of the triplets is done based on a distance computed between an anchor image and different pairs of image classes through the hierarchical tree. In addition to the methods that aim to select triplets, there are also several works that focus on reformulating the triplet loss function to emphasize the effect of hard triplets \cite{beyond-binary,triplet-focal-loss,Wang:2019}. As an example, Zhang et al. adapt the focal loss that is initially defined for classification problems and propose an extended version for triplets as an alternative to the triplet loss \cite{triplet-focal-loss}. This loss function ensures that more importance is given to hard triplets than easier ones, and thus the model can learn from the most informative triplets and converge faster. Kim et al. developed an adapted version of the triplet loss for pose estimation \cite{beyond-binary}. This loss function preserves the distance ratios from the label space in the embedding space. In~\cite{Wang:2019}, the multi-similarity loss function is proposed to reformulate the triplet loss with a weighting strategy. By using the weighting strategy, this loss function considers the relative similarity of all positive and all negative samples in a mini-batch. In~\cite{Sohn:2016}, the multi-class N-pair loss function is proposed to generalize the triplet loss function for multiple negative images associated with an anchor. In detail, for each anchor image, one positive image and several negative images are selected as hard negatives from different negative classes. In~\cite{zhang2021triplet}, the dual-anchor triplet loss function is introduced as an extension of the triplet loss. In addition to the objectives of the triplet loss, this loss function also aims at increasing the distance between the positive and negative images for a given anchor. Wang et al. extend the concept of triplets to the whole mini-batch, where all available images are first sorted and then divided into a positive set and a negative set \cite{ranked-list-loss}. Afterward, an extension of the triplet loss is used to force a margin between the two sets by using all the images. This loss function employs a weighting strategy to increase the importance of the hard negative images. In~\cite{wu2017sampling}, it is shown that when an accurate sampling strategy is considered, deep learning (DL) models with different modified loss functions provide similar accuracies. This proves the fact that triplet selection is as important as loss function in the framework of DML. Most of the triplet-based methods in CV assume that a single label is associated with each image. However, RS images typically consist of multiple classes and are associated with multi-label, which makes selecting triplets more complex than the single-label scenario.

\section{Proposed Method}
\label{sec:proposed}

\subsection{Problem Formulation}
Let $\mathbfcal{X}=\{X_1,\dots,X_M\}$ be an archive consisting of $M$ images, where $X_m$ is the $m$-th image in the archive. We assume that a training set $\mathbfcal{X}_T\subset\mathbfcal{X}$ is available. Each image in $\mathbfcal{X}_T$ is annotated with a set of class labels, which describe the content of the image. Let $\mathbb{L}=\{1,2,...,{N}\}$ be the set of all possible class labels. Each image $X_j\in\mathbfcal{X}_T$ is associated with a multi-label vector $L_j=\{l_j^1, l_j^2,...,l_j^{N}\}$, where $l_j^i=1$, if the class label $i\in\mathbb{L}$ is associated to the image $X_j$, and $l_j^i=0$ otherwise. Each training image $X_j$ is annotated with at least one class label.

\begin{figure}[t]
  \centering
  \includegraphics[width=\linewidth]{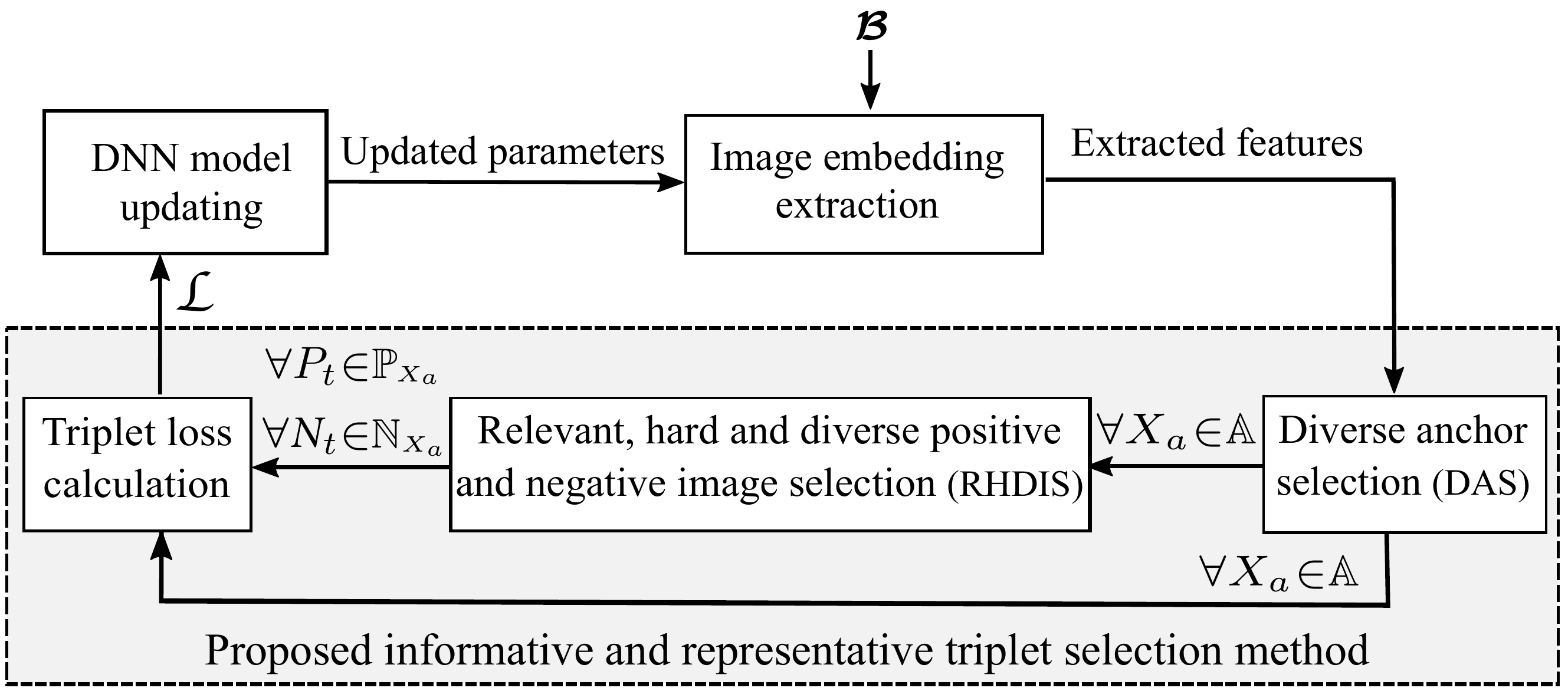}
  \caption{A block scheme of the proposed triplet sampling method to drive the training phase of a DNN for multi-label CBIR problems.}
  \label{fig:full_process}
\end{figure}

We propose a novel triplet sampling method in the framework of DL-based multi-label CBIR. The proposed method aims: i) to select a small set of informative as well as representative triplets from each training mini-batch $\mathbfcal{B}$; and ii) to accurately describe the complex semantic content of RS images. To this end, it consists of two consecutive steps: 1) selection of anchors that are diverse to each other in the feature space; 2) selection of positive and negative images with respect to each selected anchor. To achieve the latter step, we jointly evaluate the relevancy, hardness and diversity of the images during the selection (See Fig. \ref{fig:full_process}). The proposed method is independent of the considered DL model and can be used with any DL model designed for CBIR problems. In the following subsections, the two steps of the proposed method are described in detail.

\subsection{Diverse Anchor Selection (DAS)}
\label{sec:diverse-anchors}
The first step of the proposed method aims to find a small set of the most representative anchors. As mentioned before, all samples (i.e., images) in the mini-batch $\mathbfcal{B}$ could be selected as anchors. However, such an approach results in a large and redundant set of triplets and increases the computational complexity of the training. In detail, the complexity of the training grows cubically, if all possible triplets are exploited \cite{triplet-focal-loss}. Selecting a small set of anchors can significantly reduce the computational complexity of the training. To this end, we introduce a simple but efficient diverse anchor selection (DAS) strategy. The DAS strategy aims to select diverse anchors from the mini-batch that, when included in the set of triplets, can improve the retrieval performance.
To this end, it exhibits an iterative algorithm to evaluate the diversity in the feature space among the samples from the mini-batch. The algorithm starts with an empty set $\mathbb{A}=\emptyset$. The first anchor is selected randomly from the current mini-batch $\mathbfcal{B}$ and added into $\mathbb{A}$. At each iteration, a new anchor that is associated with the highest distance from all already selected anchors is selected from $\mathbfcal{B}$. In detail, at the $h$-th iteration $h$-th anchor image $X_h$ is selected as:
\begin{equation}
    X_{h}=\argmax\limits_{X_b\in\mathbfcal{B} \setminus \mathbb{A}} \left[\max\limits_{ X_a\in \mathbb{A}} {D}(X_b, X_a)\right],
\end{equation}
where $D(\cdot,\cdot)$ is the feature similarity measure, defined as the Euclidean distance between two images in the feature space. It is worth noting that the Euclidean distances are normalized based on min-max normalization. The steps are iterated until $H$ anchors are selected. Due to the selection of anchors that are as distant as possible to each other in the feature space, the diversity among the selected anchors with respect to their correlation in the feature space is maximized. This results in selecting a representative set of anchors, forming the basis for the positive and negative image selection step.
\subsection{Relevant, Hard and Diverse Positive-Negative Image Selection (RHDIS)}
\label{sec:proposed-images}
The second step of the proposed method aims to select positive and negative images for each anchor that are informative (i.e., relevant and hard) and representative (i.e., diverse to each other in the feature space). This is achieved by a novel relevant, hard and diverse positive and negative image selection strategy (RHDIS). The relevancy of an image to an anchor is defined based on its multi-label similarity with respect to the considered anchor. In detail, a positive image can be associated with high relevancy to an anchor if their class label similarity is high and vice versa. A negative image can be relevant to an anchor if its class label similarity is small and vice versa. The hardness of an image is associated with its distance to the considered anchor in the feature space. In detail, a positive image can be hard if its distance to the anchor in the embedding space is high, whereas a negative image can be considered hard if its distance to the anchor is small. 

The proposed RHDIS strategy initially evaluates the informativeness (i.e., relevancy and hardness) of the images to select the candidates for positive and negative images related to each anchor image. Then, the representative (diverse) ones among the most informative positive and negative images are selected to construct the triplets. To this end, for each image $X_b$ in the mini-batch $\mathbfcal{B}$, informativeness scores $I_p(X_a,X_b)$ (which shows if $X_b$ is a candidate positive image) and $I_n(X_a,X_b)$ (which shows if $X_b$ is a candidate negative image) with respect to anchor $X_a$ are initially computed as: 
\begin{equation}
    I_p(X_a,X_b)\equal\beta\product S(X_a,X_b)\plus(1\minus\beta)\product D(X_a,X_b),
\end{equation}
\begin{equation}
    I_n(X_a,X_b)\equal\beta\product\!\left[1 \minus S(X_a,X_b)\right]\!\plus(1\minus\beta)\product\!\left[1\minus D(X_a,X_b)\right],
\end{equation}
where $S(X_a,X_b)$ shows the class label similarity between the image $X_b$ and $X_a$. $S(X_a,X_b)\in[0,1]$ is calculated based on the soft pair-wise similarity measure (i.e., the distance between the multi-label vector $L_a$ of $X_a$ and $L_b$ of $X_b$)~\cite{soft-pairwise-similarity}. If $S(X_a,X_b)$ is high, $X_b$ can be considered as a relevant positive image, whereas if [1-$S(X_a,X_b)$] is high, $X_b$ can be considered as a relevant negative image. $D(X_a,X_b)$ is the distance between $X_b$ and $X_a$ in the embedding space and measures the hardness of images as mentioned before. If both $D(X_a,X_b)$ and $S(X_a,X_b)$ are high, the image $X_b$ can be considered as a relevant and hard positive image. If both [1-$S(X_a,X_b)$] and [1-$D(X_a,X_b)$] are high, the image $X_b$ can be considered as a relevant and hard negative image. $\beta\in[0,1]$ is the weighting parameter and can be adjusted to give more importance to either the relevancy or the hardness of the image.
 
To construct a set $\mathbb{P}_{X_a}=\{P_1,P_2,...,P_C\}$ of $C$ positive images for an anchor $X_a$, the image in the mini-batch associated with the highest $I_p$ score with respect to $X_a$ is chosen as the first positive image. Then, the next images are iteratively selected. We apply an iterative approach similar to the DAS introduced in the first step to select the most representative images. At $t$-th iteration, $t$-th positive image $P_t$ is selected as:
\begin{equation}
P_t\equal\!\!\argmax\limits_{X_b\in\mathbfcal{B}\setminus \mathbb{P}_{X_a}}\!\!\Big[\gamma\product I_p(X_a,X_b)\plus (1\minus\gamma)\product\!\!\max\limits_{ P_c\in \mathbb{P}_{X_a}}\!\!{D}(X_b, P_c)\Big].
\end{equation}
This process is repeated until the desired number of positive images is selected. The parameter $\gamma\in[0,1]$ controls the influence of the diversity term. 

To construct a set $\mathbb{N}_{X_a}=\{N_1,N_2,...,N_C\}$ of $C$ negative images for each anchor $X_a$, the image with the highest $I_n$ score in the mini-batch with regard to $X_a$ is selected as the first negative image. Afterward, the subsequent negative images are iteratively selected. At $t$-th iteration, the $t$-th negative image $N_t$ is selected as:
\begin{equation}
\!\!N_t\equal\!\!\!\argmax\limits_{X_b\in\mathbfcal{B}\setminus \mathbb{N}_{X_a}}\!\!\Big[\gamma\product I_n(X_a,X_b)\plus (1\minus\gamma)\product\!\!\!\max\limits_{ N_c\in \mathbb{N}_{X_a}}\!\!{D}(X_b, N_c)\Big].
\end{equation}
This selection strategy ensures that the selected positive and negative images for each anchor are informative (i.e., hard and relevant) and representative (i.e., diverse among each other in the feature space). After selecting the final set of triplets from the mini-batch $\mathbfcal{B}$, the triplet loss function is calculated as: 
\begin{equation}
    \mathcal{L}=\sum\limits_{\substack{\forall X_a \in \mathbb{A}\\\forall P_t \in \mathbb{P}_{X_a}\\\forall N_t \in \mathbb{N}_{X_a}}}\max\Big([D(X_a, P_t) - D(X_a, N_t) + \alpha], 0\Big),
    \label{formula:triplet-loss}
\end{equation}
\noindent where $\alpha$ is a margin enforced between positive and negative images for an anchor image. After an end-to-end training of the whole neural network by minimizing the triplet loss and learning the network parameters, the descriptors (i.e., features) of the images in $\mathbfcal{X}\setminus\mathbfcal{X}_T$ are obtained. Then, the $k$ most semantically similar images with regard to a given query image $X_q\in\mathbfcal{X}$ are selected by comparing their descriptors based on the $k$-nn algorithm.

\section{Dataset Description and Design of Experiments}
\label{sec:setup}
\subsection{Dataset Description}
\begin{figure}[t]
    \newcommand{\figwidth}{0.18\linewidth} 
    \newcommand{\figheight}{0.63in} 
    \renewcommand{\fboxsep}{0pt}
    \centering
     \begin{minipage}[t]{\figwidth}
        \centering
        \centerline{\includegraphics[height=\figheight]{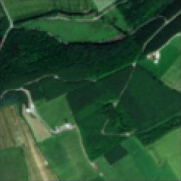}}
        \centerline{(a)}
    \end{minipage}
    \hspace{0.18in}
     \begin{minipage}[t]{\figwidth}
        \centering
        \centerline{\includegraphics[height=\figheight]{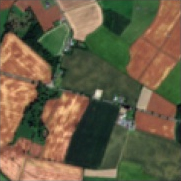}}
        \centerline{(b)}
    \end{minipage}
    \hspace{0.18in}
     \begin{minipage}[t]{\figwidth}
        \centering
        \centerline{\includegraphics[height=\figheight]{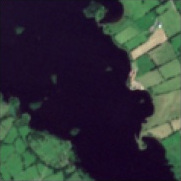}}
        \centerline{(c)}
    \end{minipage}
    \hspace{0.18in}
     \begin{minipage}[t]{\figwidth}
        \centering
        \centerline{\includegraphics[height=\figheight]{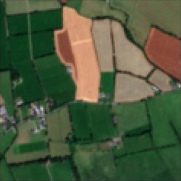}}
        \centerline{(d)}
    \end{minipage}
    \caption{An example of images from the IRS-BigEarthNet archive and their multi-labels: (a) \textit{Arable land, Pastures, Coniferous forest}, (b) \textit{Arable land, Pastures}, (c) \textit{Pastures, Inland waters}, (d) \textit{Arable land, Pastures, Complex cultivation patterns}.}
    \label{fig:ben-examples}
\end{figure}

To evaluate the proposed method, we conducted experiments on two different multi-label RS archives. The first archive is BigEarthNet \cite{bigearthnet, bigearthnet-19}, which is a large-scale multi-label Sentinel-2 benchmark archive consisting of $590,326$ images. In the experiments, we considered the images acquired over Ireland in the summer of 2017 (denoted as IRS-BigEarthNet). IRS-BigEarthNet contains $15,894$ images, each of which is made up of $120\times 120$ pixels for $10$ meter bands, $60\times 60$ pixels for $20$ meter bands and $20\times 20$ pixels for $60$ meter bands. In the experiments, we excluded the $60$ meter bands and applied bicubic interpolation to $20$ meter bands that results in 10 bands, each of which has a size of $120\times 120$ pixels. The class labels of the images were obtained from the CORINE Land Cover database of the year 2018 (CLC 2018). In the experiments, we used the 19 class nomenclature presented in~\cite{bigearthnet-19}.
As suggested in \cite{bigearthnet-19}, images with snow cover, cloud cover and cloud shadows are excluded from training and evaluation. Fig. \ref{fig:ben-examples} shows an example of images from the IRS-BigEarthNet with the associated multi-label annotations.

\begin{figure}[t]
    \newcommand{\figwidth}{0.18\linewidth}
    \newcommand{\figheight}{0.63in}
    \renewcommand{\fboxsep}{0pt}
    \centering
     \begin{minipage}[t]{\figwidth}
        \centering
        \centerline{\includegraphics[height=\figheight]{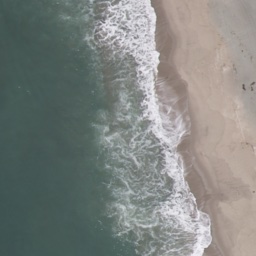}}
        \centerline{(a)}
    \end{minipage}
    \hspace{0.18in}
     \begin{minipage}[t]{\figwidth}
        \centering
        \centerline{\includegraphics[height=\figheight]{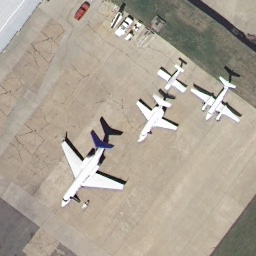}}
        \centerline{(b)}
    \end{minipage}
    \hspace{0.18in}
     \begin{minipage}[t]{\figwidth}
        \centering
        \centerline{\includegraphics[height=\figheight]{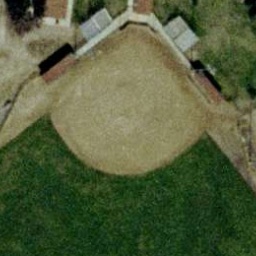}}
        \centerline{(c)}
    \end{minipage}
    \hspace{0.18in}
     \begin{minipage}[t]{\figwidth}
        \centering
        \centerline{\includegraphics[height=\figheight]{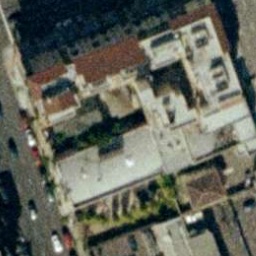}}
        \centerline{(d)}
    \end{minipage}
    \caption{An example of images from the UCMerced Land Use archive and the multi-labels associated with them: (a) \textit{sand, sea} (b) \textit{airplane, cars, grass, pavement} (c) \textit{bare-soil, buildings, grass} (d) \textit{buildings, cars, pavement, trees.}}
    \label{fig:ucm-examples}
\end{figure}
The second benchmark archive is the UC Merced Land Use (UCMerced) archive~\cite{ucmerced}, which consists of $2100$ images selected from aerial orthoimagery with a spatial resolution of 30cm. Each image has a size of $256\times 256$ pixels. The images are annotated with multi-labels by Chaudhuri et al.~\cite{ucmerced-multi}. There are 17 classes in total, with at least one and a maximum of seven class labels per image. Fig. \ref{fig:ucm-examples} shows an example of images from this archive along with their multi-label annotations.

The two benchmark archives differ greatly in size, complexity and characteristics. This allows us to demonstrate the general applicability and success of the proposed triplet sampling method in different scenarios. We randomly split UCMerced images into 60\% for training, 20\% for validation and 20\% for testing. For IRS-BigEarthNet, the officially provided splits into training, validation and evaluation sets were used. During the training step, all triplets were sampled from the training set. Query images were taken from the validation set, while image retrieval was applied to the evaluation set. 
\subsection{Design of Experiments}
In the experiments, different CNN architectures were considered as backbones, while an additional fully connected layer was added to produce image embeddings. The resulting CNNs were trained for image retrieval by means of the triplet loss. It is worth noting that our method does not depend on a specific DL model architecture. In our experiments, we evaluated three different CNN architectures: i) the shallow convolutional neural network (S-CNN) \cite{bigearthnet}; ii) DenseNet-121 \cite{densenet}; and iii) ResNet-50 \cite{resnet}. The last two architectures are well-known deep models, while the first architecture is an explicitly shallow model. All models were used without pre-training. The size of mini-batch for IRS-BigEarthNet and UCMerced was selected as $300$ and $100$, respectively. The training was performed for $100$ epochs with the Adam optimizer, using an initial learning rate of $0.001$ (which was exponentially decayed every $5$ epochs by $5$\%). The margin parameter $\alpha$ of the triplet loss was set to $0.2$. The values of $\beta$ and $\gamma$ were set to $0.5$ and $0.1$, respectively, based on a grid search strategy. All the experiments were conducted on NVIDIA Tesla V100 GPUs with $32$ GBs of memory. The results were provided in terms of the different evaluation metrics as: accuracy, precision, recall and $F_1$ score \cite{ucmerced-multi}. These values were the average of the values obtained by retrieving the $30$ and $10$ most similar images for IRS-BigEarthNet and UCMerced, respectively.

We carried out different kinds of experiments in order to: 1) perform a sensitivity analysis with respect to different network architectures and embedding sizes; 2) conduct an ablation study of the proposed triplet sampling method; 3) compare our method with different triplet sampling methods; and 4) compare our method with state of the art DML based methods. To perform the ablation study, we compared the proposed diverse anchor selection (DAS) strategy (see Section \ref{sec:diverse-anchors} for the details) with two frequently used anchor selection strategies that are:
\begin{itemize}
    \item Batch anchor selection (BAS): This strategy selects each image in the mini-batch as an anchor once and can be considered an upper bound strategy for the triplet selection. This strategy does not miss any information provided by specific triplets. However, it leads to a very high number of final triplets that can be redundant.
    \item Random anchor selection (RAS): This strategy selects a fixed number of anchors from the mini-batch without any prior assumption. It is simple, but there is no guarantee that the randomly chosen anchors provide a good basis for the triplets. 
    \end{itemize}
In the experiments, 10\% of all possible anchors from the mini-batch was chosen for the RAS and the proposed DAS strategies. We compared the proposed relevant, hard and diverse positive-negative image selection (RHDIS) strategy (see Section \ref{sec:proposed-images} for the details) with two baselines that are:
\begin{itemize}
    \item Batch positive and negative image selection (BIS): This strategy uses all images in the mini-batch. Each image is used as the positive and the negative images once. It covers all possible triplets, leading to a very high number of final triplets.
    \item Random positive and negative image selection (RIS): This strategy randomly selects sets of positive and negative images and combines all of them into triplets. Many of the resulting triplets may be trivial, but it requires no prior knowledge and provides a lower bound baseline. 
\end{itemize}
In the experiments, we also assessed the effectiveness of the joint use of the above-mentioned strategies with proposed DAS and RHDIS for the selection of anchors as well as positive and negative images. This is important as the anchor selection step is independent from the step of the positive and negative image selection, and thus proposed selection strategies can be combined with the other well-known strategies.

In the experiments, we also compared the proposed DAS-RHDIS method with two triplet sampling methods: 1) the deep metric learning using triplet network, which uses RAS for anchor selection and RIS for positive and negative image selection (denoted as TNDML)~\cite{Hoffer:2015}; and 2) enhancing remote sensing image retrieval using a triplet deep metric learning network, which employs BAS for the anchor selection and BIS for positive and negative image selection (denoted as RSDML)~\cite{cao2020enhancing}. We also compared the proposed DAS-RHDIS method with state-of-the-art DML methods for CBIR: 1) the content-based medical image retrieval (CBMIR) system, which utilizes a pair-wise similarity loss function to force all positive images to be close, while separating all the negative images with a fixed distance~\cite{Deepak:2020}; 2) the multi-similarity loss with general pair weighting for deep metric learning (denoted as MSL)~\cite{Wang:2019}; 3) the dual-anchor triplet loss (denoted as DATL) proposed in~\cite{zhang2021triplet}; and 4) the improved deep metric learning with multi-class N-pair loss objective (denoted as NPL)~\cite{Sohn:2016}. For all the methods, we used the same CNN architecture and training setup as in our method.
\begin{table}[t]
    \renewcommand{\arraystretch}{1.1}
    \setlength{\tabcolsep}{7pt}
    \centering
    \small
    \caption{The Performance of Different DL Model Architectures for the UCMerced Archive.}
    \label{tab:other-models}
    \begin{tabular}{@{}lccccc@{}}
        \toprule
        \multirow{2}{*}{\textbf{Architecture}} & \multicolumn{4}{c}{\textbf{Metric (\%)}} \\\Cline{0.75pt}{2-5}
        & Accuracy & Precision & Recall & $F_1$ Score\tabularnewline
        \toprule
        S-CNN & 40.5 & 48.9 & 51.9 & 50.3 \\
        DenseNet-121 & 45.5 & 54.4 & 58.0 & 56.1 \\
        ResNet-50 & \textbf{54.5} & \textbf{63.3} & \textbf{66.5} & \textbf{64.8} \\
        \midrule
        \bottomrule
    \end{tabular}
 \end{table} 
\section{Experimental Results}
\label{sec:results}
\subsection{Sensitivity Analysis of the Proposed Method}
In this sub-section, we present the results of the sensitivity analysis for the proposed triplet sampling method (denoted as DAS-RHDIS) in terms of different DL model architectures and different embedding sizes. To analyze the proposed DAS-RHDIS method in the framework of different DL models designed for multi-label RS CBIR, we selected the CNN architectures of: i) S-CNN; ii) DenseNet-121; and iii) ResNet-50. The embedding size for each architecture was set to $256$. In Table~\ref{tab:other-models}, the results are shown for the UCMerced archive. By assessing the table, one can observe that all the considered DL model architectures provide a high performance. As an example, although S-CNN is an explicitly shallow architecture, it achieves more than $50$\% $F_1$ score as in Dense-Net-121 and ResNet-50. This shows that the proposed DAS-RHDIS method is architecture-independent. One can also see from the table that the best scores under all metrics were obtained when ResNet-50 was utilized. As an example, ResNet-50 provides almost $9$\% higher precision and $8.5$\% higher recall compared to DenseNet-121. When compared with S-CNN, ResNet-50 leads to more than $14$\% higher $F_1$ score and accuracy. These results show that a proper selection of a DL model architecture can improve performance. For the rest of the experiments, we provided the results obtained with ResNet-50 due to its proven success.
\begin{table}[t]
    \renewcommand{\arraystretch}{1.1}
    \setlength{\tabcolsep}{10pt}
    \centering
    \small
    \caption{The Effect of Varying Embedding Sizes on the Retrieval Performance for the UCMerced Archive.}
    \label{tab:embedding-sizes}
    \begin{tabular}{@{}lcccc@{}}
        \toprule
        \multirow{2}{*}{\parbox{1.5cm}{\textbf{Embedding Size}}} & \multicolumn{4}{c}{\textbf{Metric (\%)}} \\\Cline{0.75pt}{2-5}
         & Accuracy & Precision & Recall & $F_1$ Score \tabularnewline
        \toprule
        256 & 54.5 & 63.3 & 66.5 & 64.8 \\
        512 & 56.2 & 64.6 & 69.0 & 66.7 \\
        1024 & \textbf{56.8} & \textbf{65.3} & \textbf{70.0} & \textbf{67.5} \\
        2048 & 50.3 & 58.4 & 62.8 & 60.5 \\
        \midrule
        \bottomrule
    \end{tabular}
\end{table}

In Table \ref{tab:embedding-sizes}, the results obtained by using different embedding sizes are shown for the UCMerced archive. We evaluated the effect of the embedding sizes $256$, $512$, $1024$ and $2048$ used in the proposed DAS-RHDIS method. From the table, one can see that the highest scores under all metrics are obtained when the embedding size is $1024$. Further increase of the embedding size to $2048$ does not improve the performance. As an example, the proposed method with the embedding size of $1024$ provides a $7$\% higher $F_1$ score compared to that of $2048$. This is in line with the works in literature, which demonstrate that beyond a certain size, adding any new embedding dimension may not improve the performance~\cite{xuan2018deep,oh2016deep,chen2021deep}. By analyzing the table, one can also observe that the lowest performance is obtained when the embedding size is $256$. In this case, the $F_1$ score is reduced by almost $3$\% compared to the embedding size of $1024$. Accordingly, for the rest of the experiments, we set the embedding size to $1024$. These results were also confirmed through experiments obtained by using the IRS-BigEarthNet archive (not reported for space constraints).

\subsection{Ablation Study}
In this sub-section, we performed an ablation study to analyze the effectiveness of the proposed DAS and RHDIS strategies. To demonstrate the effectiveness of the proposed DAS strategy, we compare it with RAS and BAS strategies. Table \ref{tab:results-ablation-anch} shows the results associated with the different anchor strategies for the UCMerced archive when the proposed RHDIS strategy is used for positive and negative image selection. By analyzing the table, one can observe that the proposed DAS strategy provides the highest scores under all the metrics compared to RAS and BAS. As an example, the proposed DAS strategy provides more than $7$\% higher accuracy compared to RAS under the same number of anchors (which is set to 10 in the experiments) when the positive and negative selection strategy is set to proposed RHDIS. In addition, the proposed DAS strategy leads to almost $4$\% higher recall with a smaller number of anchors compared to BAS. It is worth noting that BAS uses all the possible anchors from the mini-batch (i.e., 100 anchors). This shows the success of the proposed DAS strategy to select diverse and representative anchors with respect to random sampling and batch selection strategies.

In order to demonstrate the effectiveness of the proposed RHDIS strategy, we compare it with RIS and BIS strategies. Table \ref{tab:results-ablation-pos-neg} shows the results associated with the different positive and negative image selection strategies for the UCMerced archive when the proposed DAS strategy is used for anchor selection. From the table, one can see that the proposed RHDIS strategy achieves the highest performance under all metrics compared to RIS and BIS. As an example, the recall of the proposed RHDIS strategy is more than $8$\% higher compared to that of BIS when the anchor selection strategy is set to proposed DAS. It is worth noting that BIS exploits all positive and negative images in the batch, while RHDIS relies on a much smaller number of triplets to achieve this result. The performance of RIS is lower than RHDIS and BIS under each metric when the anchor selection strategy is set to proposed DAS. For example, the recall obtained by RIS is about $10$\% lower than that of proposed RHDIS under the same number of triplets. This shows the effectiveness of the proposed RHDIS selection strategy to select relevant, hard and diverse positive-negative images compared to random sampling and batch selection strategies for a given set of anchors. These results were also confirmed through experiments obtained by using the IRS-BigEarthNet archive.

\begin{table}[t]
    \renewcommand{\arraystretch}{1.15}
    \setlength{\tabcolsep}{4.5pt}
    \centering
    \small
    \caption{Results obtained by the different anchor selection strategies (RAS, BAS and proposed DAS) under different metrics for the UCMerced archive when proposed RHDIS is used for positive and negative image selection.}
    \label{tab:results-ablation-anch}
    \begin{tabular}{@{}lcccc@{}}
        \toprule
        \multirow{2}{*}{\parbox{2.8cm}{\textbf{Anchor\\Selection Strategy}}} & \multicolumn{4}{c}{\textbf{Metric (\%)}} \\\Cline{0.75pt}{2-5}
        & Accuracy & Precision & Recall & $F_1$ Score \\
        \toprule
        RAS & 49.2 & 58.1 & 61.9 & 60.0 \\
        BAS & 53.5 & 62.0 & 66.5 & 64.2 \\
        Proposed DAS & \textbf{56.8} & \textbf{65.3} & \textbf{70.0} & \textbf{67.5} \\
        \midrule
        \bottomrule
    \end{tabular}
\end{table}

\begin{table}[t]
    \renewcommand{\arraystretch}{1.15}
    \setlength{\tabcolsep}{4.5pt}
    \centering
    \small
    \caption{Results obtained by the different positive and negative image selection strategies (RIS, BIS and proposed RHDIS) under different metrics for the UCMerced archive when proposed DAS is used for anchor selection.}
    \label{tab:results-ablation-pos-neg}
    \begin{tabular}{@{}lcccc@{}}
        \toprule
        \multirow{2}{*}{\parbox{2.9cm}{\textbf{Positive and Negative Selection Strategy}}} & \multicolumn{4}{c}{\textbf{Metric (\%)}} \\\Cline{0.75pt}{2-5}
        & Accuracy & Precision & Recall & $F_1$ Score \\
        \toprule
        RIS & 48.6 & 57.4 & 60.1 & 58.7 \\
        BIS & 48.9 & 57.6 & 61.4 & 59.4
        \\
        Proposed RHDIS & \textbf{56.8} & \textbf{65.3} & \textbf{70.0} & \textbf{67.5} \\
        \midrule
        \bottomrule
    \end{tabular}
\end{table}

\begin{table*}[t]
    \renewcommand{\arraystretch}{1.2}
    \setlength{\tabcolsep}{10pt}
    \centering
    \small
    \caption{The performance of different triplet selection methods for the IRS-BigEarthNet and UCMerced archives.}
    \label{tab:sota-triplet}
    \begin{tabular}{@{}llcccc@{}}
        \Cline{1pt}{1-6}
        \multirow{2}{*}{\textbf{Archive}} & \multirow{2}{*}{\textbf{Method}} & \multicolumn{4}{c}{\textbf{Metric (\%)}} \\\Cline{0.75pt}{3-6}
        & & Accuracy & Precision & Recall & $F_1$ Score \\\Cline{1pt}{1-6}
        \multirow{3}{*}{IRS-BigEarthNet}
        & TNDML~\cite{Hoffer:2015} & 59.3 & 73.7 & 73.8 & 73.8 \\\Cline{0.5pt}{2-6}
        & RSDML~\cite{cao2020enhancing} & 60.2 & 75.4 & 73.9 & 74.6 \\\Cline{0.5pt}{2-6}
        & Proposed DAS-RHDIS & \textbf{62.7} & \textbf{77.7} & \textbf{75.7} & \textbf{76.7} \\\Cline{1pt}{1-6}
        
        \multirow{3}{*}{UCMerced}
        & TNDML~\cite{Hoffer:2015} & 44.0 & 52.6 & 55.8 & 54.2 \\\Cline{0.5pt}{2-6}
        & RSDML~\cite{cao2020enhancing} & 48.4 & 56.3 & 61.9 & 59.0  \\\Cline{0.5pt}{2-6}
        & Proposed DAS-RHDIS & \textbf{56.8} & \textbf{65.3} & \textbf{70.0} & \textbf{67.5} \\
        \Cline{1pt}{1-6}
    \end{tabular}
\end{table*}

\subsection{Comparison of the Proposed Method with Different Triplet Sampling Methods}

In this sub-section, we evaluate the effectiveness of the proposed DAS-RHDIS method compared to different triplet selection methods, which are: TNDML~\cite{Hoffer:2015}, and RSDML~\cite{cao2020enhancing}. Table~\ref{tab:sota-triplet} shows the corresponding image retrieval performances on the IRS-BigEarthNet and the UCMerced archives. By analyzing the table, one can see that the proposed DAS-RHDIS method leads to the highest scores under all metrics for both archives. For example, DAS-RHDIS outperforms TNDML by $4$\% in precision and more than $3$\% in accuracy for the IRS-BigEarthNet archive, more than $13$\% in $F_1$ score and almost $15$\% in recall for the UCMerced archive. The proposed DAS-RHDIS method provides about $2$\% higher and $8$\% higher $F_1$ scores compared to the RSDML method for IRS-BigEarthNet and UCMerced, respectively. These results demonstrate the success of the proposed DAS-RHDIS method compared to other triplet sampling methods.

\begin{figure}[t!]
    \newcommand{\figwidth}{0.18\linewidth} 
    \newcommand{\figheight}{0.63in} 
    \renewcommand{\fboxsep}{0pt}
    \newcommand{\fig}[2]{images/fig6/#1_ben_#2.png}
    \centering
     \begin{minipage}[t]{\figwidth}
        \centering
        \centerline{\includegraphics[height=\figheight]{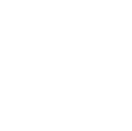}}
    \end{minipage}  
     \begin{minipage}[t]{\figwidth}
        \centering
        \centerline{\includegraphics[height=\figheight]{images/fig6/empty.png}}
    \end{minipage}
     \begin{minipage}[t]{\figwidth}
        \centering
        \centerline{\includegraphics[height=\figheight]{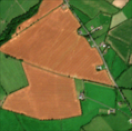}}
        \vspace{-0.05in}\centerline{(a)}\medskip\vspace{-0.05in}
    \end{minipage}
     \begin{minipage}[t]{\figwidth}
        \centering
        \centerline{\includegraphics[height=\figheight]{images/fig6/empty.png}}
    \end{minipage}
     \begin{minipage}[t]{\figwidth}
        \centering
        \centerline{\includegraphics[height=\figheight]{images/fig6/empty.png}}
    \end{minipage}
          \begin{minipage}[t]{\figwidth}
        \centering
        \centerline{2\textsuperscript{nd}}\medskip\vspace{-0.05in}
        \centerline{\includegraphics[height=\figheight]{\fig{rasris}{2}}}
    \end{minipage}    
          \begin{minipage}[t]{\figwidth}
        \centering
        \centerline{5\textsuperscript{th}}\medskip\vspace{-0.05in}
        \centerline{\includegraphics[height=\figheight]{\fig{rasris}{5}}}
    \end{minipage}
          \begin{minipage}[t]{\figwidth}
        \centering
        \centerline{10\textsuperscript{th}}\medskip\vspace{-0.05in}
        \centerline{\includegraphics[height=\figheight]{\fig{rasris}{10}}}
        \vspace{-0.05in}\centerline{(b)}\medskip\vspace{-0.05in}
    \end{minipage}
          \begin{minipage}[t]{\figwidth}
        \centering
        \centerline{15\textsuperscript{th}}\medskip\vspace{-0.05in}
        \centerline{\includegraphics[height=\figheight]{\fig{rasris}{15}}}
    \end{minipage}
\begin{minipage}[t]{\figwidth}
        \centering
        \centerline{20\textsuperscript{th}}\medskip\vspace{-0.05in}
        \centerline{\includegraphics[height=\figheight]{\fig{rasris}{20}}}
    \end{minipage}
          \begin{minipage}[t]{\figwidth}
        \centering
        \centerline{\includegraphics[height=\figheight]{\fig{basbis}{2}}}
    \end{minipage}
          \begin{minipage}[t]{\figwidth}
        \centering
        \centerline{\includegraphics[height=\figheight]{\fig{basbis}{5}}}
    \end{minipage}
          \begin{minipage}[t]{\figwidth}
        \centering
        \centerline{\includegraphics[height=\figheight]{\fig{basbis}{10}}}
        \vspace{-0.05in}\centerline{(c)}\medskip\vspace{-0.05in}
    \end{minipage}
          \begin{minipage}[t]{\figwidth}
        \centering
        \centerline{\includegraphics[height=\figheight]{\fig{basbis}{15}}}
    \end{minipage}
          \begin{minipage}[t]{\figwidth}
        \centering
        \centerline{\includegraphics[height=\figheight]{\fig{basbis}{20}}} 
    \end{minipage}
          \begin{minipage}[t]{\figwidth}
        \centering
        \centerline{\includegraphics[height=\figheight]{\fig{dasrhdis}{2}}}
    \end{minipage}
          \begin{minipage}[t]{\figwidth}
        \centering
        \centerline{\includegraphics[height=\figheight]{\fig{dasrhdis}{5}}}
    \end{minipage}
          \begin{minipage}[t]{\figwidth}
        \centering
        \centerline{\includegraphics[height=\figheight]{\fig{dasrhdis}{10}}}
        \vspace{-0.05in}\centerline{(d)}\medskip\vspace{-0.05in}
    \end{minipage}
          \begin{minipage}[t]{\figwidth}
        \centering
        \centerline{\includegraphics[height=\figheight]{\fig{dasrhdis}{15}}}
    \end{minipage}
          \begin{minipage}[t]{\figwidth}
        \centering
        \centerline{\includegraphics[height=\figheight]{\fig{dasrhdis}{20}}}
    \end{minipage}
    \caption{An image retrieval example: (a) query image; (b) images retrieved by TNDML; (c) images retrieved by RSDML; (d) images retrieved by the proposed DAS-RHDIS method (IRS-BigEarthNet archive).}
    \label{fig:retrieval-result}
\end{figure}
\begin{figure}[t]
    \newcommand{\figwidth}{0.18\linewidth} 
    \newcommand{\figheight}{0.63in} 
    \renewcommand{\fboxsep}{0pt}
    \newcommand{\fig}[2]{images/fig7/#1_ucm_#2.png}
    \centering
     \begin{minipage}[t]{\figwidth}
        \centering
        \centerline{\includegraphics[height=\figheight]{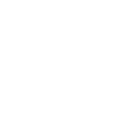}}
    \end{minipage}  
     \begin{minipage}[t]{\figwidth}
        \centering
        \centerline{\includegraphics[height=\figheight]{images/fig7/empty.png}}
    \end{minipage}
     \begin{minipage}[t]{\figwidth}
        \centering
        \centerline{\includegraphics[height=\figheight]{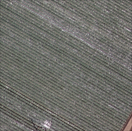}}
        \vspace{-0.05in}\centerline{(a)}\medskip\vspace{-0.05in}
    \end{minipage}
     \begin{minipage}[t]{\figwidth}
        \centering
        \centerline{\includegraphics[height=\figheight]{images/fig7/empty.png}}
    \end{minipage}
     \begin{minipage}[t]{\figwidth}
        \centering
        \centerline{\includegraphics[height=\figheight]{images/fig7/empty.png}}
    \end{minipage}
          \begin{minipage}[t]{\figwidth}
        \centering
        \centerline{2\textsuperscript{nd}}\medskip\vspace{-0.05in}
        \centerline{\includegraphics[height=\figheight]{\fig{rasris}{2}}}
    \end{minipage}    
          \begin{minipage}[t]{\figwidth}
        \centering
        \centerline{5\textsuperscript{th}}\medskip\vspace{-0.05in}
        \centerline{\includegraphics[height=\figheight]{\fig{rasris}{5}}}
    \end{minipage}
          \begin{minipage}[t]{\figwidth}
        \centering
        \centerline{10\textsuperscript{th}}\medskip\vspace{-0.05in}
        \centerline{\includegraphics[height=\figheight]{\fig{rasris}{10}}}
        \vspace{-0.05in}\centerline{(b)}\medskip\vspace{-0.05in}
    \end{minipage}
          \begin{minipage}[t]{\figwidth}
        \centering
        \centerline{15\textsuperscript{th}}\medskip\vspace{-0.05in}
        \centerline{\includegraphics[height=\figheight]{\fig{rasris}{15}}}
    \end{minipage}
\begin{minipage}[t]{\figwidth}
        \centering
        \centerline{20\textsuperscript{th}}\medskip\vspace{-0.05in}
        \centerline{\includegraphics[height=\figheight]{\fig{rasris}{20}}}
    \end{minipage}
          \begin{minipage}[t]{\figwidth}
        \centering
        \centerline{\includegraphics[height=\figheight]{\fig{basbis}{2}}}
    \end{minipage}
          \begin{minipage}[t]{\figwidth}
        \centering
        \centerline{\includegraphics[height=\figheight]{\fig{basbis}{5}}}
    \end{minipage}
          \begin{minipage}[t]{\figwidth}
        \centering
        \centerline{\includegraphics[height=\figheight]{\fig{basbis}{10}}}
        \vspace{-0.05in}\centerline{(c)}\medskip\vspace{-0.05in}
    \end{minipage}
          \begin{minipage}[t]{\figwidth}
        \centering
        \centerline{\includegraphics[height=\figheight]{\fig{basbis}{15}}}
    \end{minipage}
          \begin{minipage}[t]{\figwidth}
        \centering
        \centerline{\includegraphics[height=\figheight]{\fig{basbis}{20}}} 
    \end{minipage}
          \begin{minipage}[t]{\figwidth}
        \centering
        \centerline{\includegraphics[height=\figheight]{\fig{dasrhdis}{2}}}
    \end{minipage}
          \begin{minipage}[t]{\figwidth}
        \centering
        \centerline{\includegraphics[height=\figheight]{\fig{dasrhdis}{5}}}
    \end{minipage}
          \begin{minipage}[t]{\figwidth}
        \centering
        \centerline{\includegraphics[height=\figheight]{\fig{dasrhdis}{10}}}
        \vspace{-0.05in}\centerline{(d)}\medskip\vspace{-0.05in}
    \end{minipage}
          \begin{minipage}[t]{\figwidth}
        \centering
        \centerline{\includegraphics[height=\figheight]{\fig{dasrhdis}{15}}}
    \end{minipage}
          \begin{minipage}[t]{\figwidth}
        \centering
        \centerline{\includegraphics[height=\figheight]{\fig{dasrhdis}{20}}}
    \end{minipage}
    \caption{An image retrieval example: (a) query image; (b) images retrieved by TNDML; (c) images retrieved by RSDML; (d) images retrieved by the proposed DAS-RHDIS method (UCMerced archive).}
    \label{fig:retrieval-result-ucm}
\end{figure}

Fig. \ref{fig:retrieval-result} shows an example of images retrieved from IRS-BigEarthNet by TNDML, RSDML and the proposed DAS-RHDIS when the query image contains \textit{Arable land}, \textit{Pastures} and \textit{Complex cultivation patterns}. The retrieval order of images is given below the query image. By analyzing the figure, one can observe that the classes of \textit{Pasture} and \textit{Arable land} are very prominent in all retrieved images by RSDML and DAS-RHDIS, while TNDML provides similar images to the query only at the retrieved orders of 5 and 10. When DAS-RHDIS is compared with RSDML, the proposed method retrieves semantically more similar images. One of the reasons is that the RSDML relies only on the class label similarity, while the proposed DAS-RHDIS method: i) extracts and exploits the semantic content of the images; and ii) considers the diversity and hardness of images during triplet selection. We observed similar behavior for the UCMerced archive. Fig. \ref{fig:retrieval-result-ucm} shows an example of images retrieved from UCMerced. The query image for this example only contains the \textit{Field} class. Most of the images retrieved by the proposed method (except the 20$^{th}$ image) belong to the same class with the query (see Fig. \ref{fig:retrieval-result-ucm}-d). However, only a small number of images retrieved by the TNDML and the RSDML methods contains the \textit{Field} class (see Fig. \ref{fig:retrieval-result-ucm}-b and \ref{fig:retrieval-result-ucm}-c). 

\begin{table*}[t]
    \renewcommand{\arraystretch}{1.1}
    \setlength{\tabcolsep}{10pt}
    \centering
    \small
    \caption{The performance of different deep metric learning methods for the IRS-BigEarthNet and UCMerced archives.}
    \label{table:soa}
    \begin{tabular}{@{}llcccc@{}}
        \Cline{1pt}{1-6}
        \multirow{2}{*}{\textbf{Archive}} & \multirow{2}{*}{\textbf{Method}} & \multicolumn{4}{c}{\textbf{Metric (\%)}} \\\Cline{0.75pt}{3-6}
        & & Accuracy & Precision & Recall & $F_1$ Score \\\Cline{1pt}{1-6}
        \multirow{5}{*}{IRS-BigEarthNet}
        & CBMIR~\cite{Deepak:2020} & 59.6 & 73.2 & 74.6 & 73.9 \\\Cline{0.5pt}{2-6}
        & MSL~\cite{Wang:2019} & 57.9 & 75.0 & 68.7 & 71.7 \\\Cline{0.5pt}{2-6}
        & DATL~\cite{zhang2021triplet} & 60.6 & 75.3  & 74.0  & 74.7 \\\Cline{0.5pt}{2-6}
        & NPL~\cite{Sohn:2016} & 60.8 & 76.5 & 72.6 & 74.5 \\\Cline{0.5pt}{2-6}
        & Proposed DAS-RHDIS & \textbf{62.7} & \textbf{77.7} & \textbf{75.7} & \textbf{76.7} \\\Cline{1pt}{1-6}
        \multirow{5}{*}{UCMerced}
        & CBMIR~\cite{Deepak:2020} & 42.0 & 50.9 & 53.0 & 51.9 \\\Cline{0.5pt}{2-6}
        & MSL~\cite{Wang:2019} & 46.6 & 58.1 & 61.0 & 59.5 \\\Cline{0.5pt}{2-6}
        & DATL~\cite{zhang2021triplet} & 48.7 & 57.2 & 60.7 & 58.9 \\\Cline{0.5pt}{2-6}
        & NPL~\cite{Sohn:2016} & 51.8 & 61.5 & 58.7 & 60.1 \\\Cline{0.5pt}{2-6}
        & Proposed DAS-RHDIS & \textbf{56.8} & \textbf{65.3} & \textbf{70.0} & \textbf{67.5} \\
        \Cline{1pt}{1-6}
    \end{tabular}
\end{table*}
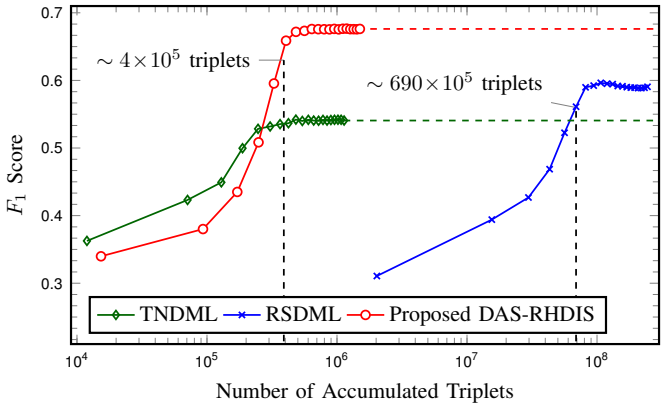
\begin{figure}[t]
    \centering
    \begin{tikzpicture}[scale = 0.83]%
	\begin{axis}[
    legend columns=3,
    height=7cm,
    width=11cm,
    legend style={font=\normalsize},
	legend pos=south west,
    xmode=log,
    minor x tick num=4,
    minor y tick num=5,
    xlabel= {\normalsize Number of Accumulated Triplets},
    ylabel= {\normalsize $F_1$ Score},
    xmin=9000,xmax=300000000,
    ymin=0.21,ymax=0.71]
    \addplot+[name path=capacity,mark=diamond,color=black!60!green,mark options={fill=white},line width=0.8pt] table [x=Step, y=Value, col sep=comma] {data/TNDML.csv};\addlegendentry{TNDML};
    \addplot+[name path=reused,mark=x,color=blue,mark options={fill=white},line width=0.8pt] table [x=Step, y=Value, col sep=comma] {data/RSDML.csv};
    \node[coordinate, pin={[fill=white, pin distance = 4 mm]-190:{$\sim690\!\times\!10^5$ triplets}}]
    at (axis cs:68500903,0.56) {};\addlegendentry{RSDML};
    \addplot+[name path=capacity,color=red,mark=*,mark options={fill=white},line width=0.8pt] table [x=Step, y=Value, col sep=comma] {data/DAS-RHDIS.csv};\addlegendentry{Proposed DAS-RHDIS};
    \node[coordinate, pin={[fill=white, pin distance = 4 mm]-180:{$\sim  4\!\times\!10^5$ triplets}}]
    at (axis cs:390000,0.63) {};
    \addplot[color=black,dashed,thick] coordinates {(390000,0) (390000,0.63)};
    \addplot[color=black,dashed,thick] coordinates {(69000903,0) (69000903,0.56)};
    \addplot[color=red,dashed, thick] coordinates {(709310,0.676151394844055) (354381356,0.676151394844055)};
    \addplot[color=black!60!green,dashed, thick] coordinates {(559310,0.540641784667969) (354381356,0.540641784667969)};
	\end{axis}
    \end{tikzpicture}%
    \caption{$F_1$ scores obtained by different triplet sampling strategies and the number of accumulated triplets during the training (The UCMerced archive).}
    \label{fig:number-triplets}
\end{figure}
During the learning of a metric space by using the triplet loss, a small subset of the available triplets carries the information needed to learn an accurate representation for image retrieval. The proposed DAS-RHDIS identifies these triplets and only learns from a subset of selected informative and representative samples, reducing the number of training triplets. Fig. \ref{fig:number-triplets} shows the performance of TNDML, RSDML and the proposed DAS-RHDIS method in terms of the number of accumulated training triplets under the same number of epochs (which is set to 100 in the experiments) for the UCMerced archive. The horizontal axis shows the number of triplets in a logarithmic scale, while the vertical axis shows the corresponding $F_1$ scores. The performance is associated with the numbers of triplets, which are utilized by the considered triplet selection method. The annotation points indicate the number of triplets needed for the considered method to reach at least $90$\% of its final performance. From the figure, one can observe that even after the last training epoch of the proposed DAS-RHDIS method, the total number of triplets is significantly smaller than the first epoch of the RSDML method. During training, the RSDML selects more triplets at each epoch compared to the other two methods. This is due to the characteristic of RSDML that selects all the possible triplets from a mini-batch, which grows cubically. The final $F_1$ score of our proposed method is more than $8$\% higher than RSDML with significantly less number of total triplets. One can also see from the figure that TNDML (which uses random triplet selection) under the same number of triplets with our method leads to a significant performance drop. The $F_1$ score obtained by TNDML is $13$\% lower than the $F_1$ score obtained by the proposed DAS-RHDIS method. These results show the effectiveness of our method to select a subset of informative triplets during training, resulting in faster convergence and a performance gain in the retrieval.

\subsection{Comparison of the Proposed Method with the State-of-the-Art DML Approaches}
In this sub-section, we assessed the effectiveness of the proposed DAS-RHDIS method compared to the state-of-the-art deep metric learning approaches, which are: CBMIR~\cite{Deepak:2020}, MSL~\cite{Wang:2019}, DATL~\cite{zhang2021triplet} and NPL~\cite{Sohn:2016}. Table~\ref{table:soa} shows the results under different metrics for the IRS-BigEarthNet and UCMerced archives. By analyzing the table, one can see that the proposed DAS-RHDIS method leads to the highest scores under all metrics for both archives. As an example, the proposed DAS-RHDIS method provides $2$\% higher and $8$\% higher accuracy compared to the DATL method for IRS-BigEarthNet and UCMerced, respectively. The table also shows that the CBMIR and the MSL methods obtain the lowest scores in most of the metrics. For example, CBMIR provides more than $4$\% lower and $14$\% lower precision than the proposed DAS-RHDIS for IRS-BigEarthNet and UCMerced, respectively. Since the loss function in CBMIR forces a fixed distance for all images, it is more restrictive compared to the triplet-based DML losses. This can lead to learning the metric space, in which the similarity between the images are not properly characterized~\cite{wu2017sampling}. When compared with the MSL method, DAS-RHDIS achieves $7$\% higher recall and more than $4$\% higher accuracy for the IRS-BigEarthNet archive, more than $7$\% higher precision and $8$\% higher $F_1$ score for the UCMerced archive. Despite the proven success of the MSL method for single label images, we observed that the full capacity of this method is not applicable for multi-label images. Since the MSL method considers all the possible negatives and positives and their relative feature distances among each other, its performance is very sensitive to the proper definition of the positive and the negative sets for a given anchor. However, the evident distinction of these sets is difficult to achieve for multi-label images. When compared with the NPL method, the proposed DAS-RHDIS method provides $2$\% higher and $7$\% higher $F_1$ scores for IRS-BigEarthNet and UCMerced, respectively. It is worth noting that NPL obtains relatively closer results to the proposed DAS-RHDIS due to its negative mining strategy. NPL uses an extension of the triplet loss, which selects multiple negative images from different negative classes for each anchor and positive image. This negative mining strategy allows NPL to include class-based diversity among the negative samples. However, in NPL, the hardness and diversity in the positive samples are not considered, resulting in the selection of trivial triplets. This can affect its performance for the retrieval task. The proposed DAS-RHDIS identifies informative and representative triplets by relying on the relevancy, hardness and diversity of images. This allows us to reach more effective image retrieval performance compared to the other methods.

\section{Conclusion}
\label{sec:conclusion}
This paper introduces a novel method to select a set of informative and representative triplets from multi-label training images to achieve deep metric learning for multi-label CBIR problems in RS. The proposed triplet sampling method is defined based on a two-steps procedure and applied on each training mini-batch of a DL-based retrieval system. In the first step, diverse anchor images are selected based on a simple but efficient iterative algorithm. Then, in the second step, sets of positive and negative images for each anchor are selected based on relevancy, hardness and diversity of the positive and negative images. Finally, the triplets are constructed from the selected anchors and their respective positive and negative images. Through the above-mentioned steps, the proposed method results in selecting a compact subset of informative and representative triplets, which enables accurate and efficient learning of DL models for multi-label CBIR in RS. Experimental results obtained on two multi-label RS benchmark archives under different DL architectures show the effectiveness of the proposed method in CBIR problems. In detail, the results have demonstrated that most of the available triplets do not contribute to the learning progress and can be safely discarded. Focusing on a small informative and representative subset is sufficient for achieving comparable performance compared to the case, for which all possible triplets are used. It is worth noting that the proposed triplet sampling method does not rely on a specific DL architecture and can be adapted to any metric learning method. 

As a final remark, we would like to point out that the proposed method currently relies on the class labels to select positive and negative images for each anchor. As a future work, we plan to develop an unsupervised strategy that can select informative positive and negative images without requiring any land-use land-cover class label.
\appendices

\section*{Acknowledgments}
This work is funded by the European Research Council (ERC) through the ERC-2017-STG BigEarth Project under Grant 759764 and by the German Research Foundation as part of the priority programme “Volunteered Geographic Information: Interpretation, Visualisation and Social Computing” (VGIscience, priority programme 1894). The Authors would like to thank Tristan Kreuziger for the valuable discussions on triplet sampling as well as for the design of Fig. 2. 

\bibliography{ref}
\bibliographystyle{IEEEtran}
\end{document}